\documentclass{article}
\usepackage{ijcai22}

\usepackage{times}
\usepackage{soul}
\usepackage{url}
\usepackage[hidelinks]{hyperref}
\usepackage[utf8]{inputenc}
\usepackage[small]{caption}
\usepackage{graphicx}
\usepackage{amsmath}
\usepackage{amsthm}
\usepackage{amsfonts}
\usepackage{booktabs}
\usepackage{algorithm}
\usepackage{algorithmic}
\usepackage{mathtools}
\usepackage[detect-weight=true, binary-units=true]{siunitx}
\usepackage{url,hyperref,cleveref}
\Crefname{equation}{Eq.}{Eqs.}
\Crefname{figure}{Fig.}{Figs.}
\Crefname{section}{Sec.}{Secs.}
\crefname{algocf}{alg.}{algs.}
\Crefname{algocf}{Algorithm}{Algorithms}
\usepackage[normalem]{ulem} 
\usepackage{color,colortbl,etoolbox,xcolor} 
\robustify\bfseries
\robustify\uline
\usepackage{comment}
\usepackage{wrapfig}
\usepackage{multirow,booktabs,makecell,array}

\title{Adults as Augmentations for Children in Facial Emotion Recognition with Contrastive Learning}

\author{
Marco Virgolin$^1$
\and
Andrea De Lorenzo$^2$\and
Tanja Alderliesten$^3$\And
Peter A.~N.~Bosman$^{1,4}$
\affiliations
$^1$Centrum Wiskunde \& Informatica\\
$^2$University of Trieste\\
$^3$Leiden University Medical Center\\
$^4$Delft University of Technology
\emails
\{marco.virgolin, peter.bosman\}@cwi.nl,
andrea.delorenzo@units.it,
t.alderliesten@lumc.nl
}

\begin{document}

\maketitle

\begin{abstract}
Emotion recognition in children can help the early identification of, and intervention on, psychological complications that arise in stressful situations such as cancer treatment.
Though deep learning models are increasingly being adopted, data scarcity is often an issue in pediatric medicine, including for facial emotion recognition in children.
In this paper, we study the application of data augmentation-based contrastive learning to overcome data scarcity in facial emotion recognition for children.
We explore the idea of ignoring generational gaps, by adding abundantly available adult data to pediatric data, to learn better representations.
We investigate different ways by which adult facial expression images can be used alongside those of children. 
In particular, we propose to explicitly incorporate within each mini-batch adult images as augmentations for children's.
Out of $84$ combinations of learning approaches and training set sizes, we find that supervised contrastive learning with the proposed training scheme performs best, reaching a test accuracy that typically surpasses the one of the second-best approach by $2\%$ to $3\%$.
Our results indicate that adult data can be considered to be a meaningful augmentation of pediatric data for the recognition of emotional facial expression in children, and open up the possibility for other applications of contrastive learning to improve pediatric care by complementing data of children with that of adults.
\end{abstract}

\section{Introduction}
The course of vital treatments such as cancer therapy can expose the patient to high levels of stress and depression, in adults as well as in children \cite{linden2012anxiety,compas2014children}.
In particular for young children, whose ability to disclose inner feelings is still under development \cite{sprung2015children}, being able to recognize non-verbal emotion signals can be a critical factor for the monitoring of, and intervention on, their well-being. 
Here we consider the problem of automating emotion recognition from images of facial expressions of children.

State-of-the-art methods to tackle facial emotion recognition are based on deep learning, which is notoriously data-hungry: tens of thousands of images may be needed to obtain good accuracy on a given task \cite{li2020deep}.
Unfortunately, in many medical applications, the available data are limited. 
This problem is typically exacerbated in the case of pediatric data. 
Reasons for the lack of pediatric data include, e.g., certain pathologies being less likely to occur in younger people \cite{li2008cancer} and regulations to protect children's privacy \cite{tikkinen2018eu}. 
For emotion recognition from facial expressions, the relatively up-to-date repository ``Awesome-FER''~\cite{fan2021awesome} lists only one children-specific data set out of the 20 reported, while the Wikipedia page ``facial expression databases''\footnote{\url{https://en.wikipedia.org/wiki/Facial_expression_databases}} lists none out of the 19 reported.

When only limited data are available, \emph{data augmentation} can be of great help. 
Typically, data augmentation on images is carried out by randomly applying modifications such as cropping, flipping, and color jittering, which, crucially, do not change the \emph{semantics} of the original image\footnote{The semantics of an image depend on the recognition problem. For example, an augmentation that flips images horizontally is obviously detrimental if the goal is to classify the direction pointed by arrow images.}\cite{shorten2019survey}. 
In this paper, we study the effect of data augmentation for facial emotion recognition in children, and in particular whether \emph{introducing adult facial expressions as augmentations for child facial expressions is better than solely relying on repeated augmentations of child facial expressions}.
The hypotheses that motivate taking this approach are, first, that the extent of information that can be learned from inflating a dataset with classic augmentations has diminishing returns; and second, that facial features that convey an emotion are orthogonal to age, hence using data for emotion recognition of adults can be considered to be but another form of augmentation for data regarding children.

To conduct this study, we rely on \emph{contrastive learning}, a recent class of methods that uses data augmentations to learn a latent representation where similar data points are close in some metric space and dissimilar ones are far (e.g., in terms of inner product).
We particularly build upon \emph{Supervised Contrastive learning} (SupCon) \cite{khosla2020supervised}, by proposing a scheme where images of adults are explicitly paired with images of children that share the same emotion.
For comparison, we include standard SupCon, the unsupervised approach \emph{Simple framework for Contrastive Learning of visual Representations} (SimCLR) \cite{chen2020simple}, and traditional supervised learning, with and without expanding the original child emotion data set with an adult one.


\section{Related work}

Contrastive learning concerns learning a function that maps data points that are similar (resp., dissimilar) into latent representations that are near (resp., far), in terms of a certain metric space (e.g., the Euclidean one). 
These latent representations can be useful to perform supervised learning tasks (if labels are available), as well as to reason about the data. 
There exist several approaches to contrastive learning, some dating back to the early the 2000s (e.g.,~\cite{chopra2005learning}), which differ in how the training process works, what loss is used, and whether the approach is supervised (i.e., labels are used) or unsupervised (no labels are used, e.g., because they are  unavailable)~\cite{jaiswal2021survey,weng2021contrastive}.
For example, the method proposed by~\cite{schroff2015facenet} for face identification proposed a \emph{triplet loss}, which trains a neural network to minimize the L2-distance between an \emph{anchor} (i.e., an image of a person) with its \emph{positives} (i.e., other images of the same person) and maximize it with its \emph{negatives} (i.e., images of other people), using labels to define what images portrait the same person.
A different and label-free (i.e., unsupervised or self-supervised) approach can typically be taken when knowing how modifications of the original data that do not alter their semantics can be generated. 
A well-known unsupervised contrastive approach for images is SimCLR~\cite{chen2020simple}, whereby augmentations are used to generate different versions of a same image that carry the same meaning (e.g., by cropping, rotating, color jittering, and so on). 
The loss function of SimCLR is designed to learn latent representations such that the latent representations of augmented images that share a common origin are similar with respect to the metric space. 
SupCon~\cite{chen2020simple} is similar to SimCLR but, besides using augmentations, also (requires and) uses the labels of the images, to learn better latent representations.

In this work, we focus on facial emotion recognition in children.
While there exist many works in literature about facial emotion recognition in general~\cite{ko2018brief}, the number of studies that specifically focus on children are limited.
Of these, recent works normally rely on deep convolutional neural networks, since these methods represent the state-of-the-art in image classification in general~\cite{rawat2017deep,anwar2018medical}; and also on data sets such as CAFE (\num{1192} images of 2 to 8 years old children)~\cite{lobue2015child}, the Dartmouth database of children's faces (640 images, 6--16 y.o.)~\cite{dalrymple2013dartmouth}, NIMH-ChEFS (482 images, 10--17 y.o.)~\cite{egger2011nimh}, and EmoReact (1102 audio-visual clips, 4--14 y.o.)~\cite{nojavanasghari2016emoreact}.
Since these data sets are relatively small for training convolutional neural networks, research works typically involve pre-training, e.g., on a larger data set of adults~\cite{lopez2019emotion}, or feature reduction, e.g., by using facial landmarks rather than the entire image as inputs for the network~\cite{rao2020emotion}.
To the best of our knowledge, this is the first paper that considers contrastive learning for facial emotion recognition in children.


\section{Methods}

\begin{figure}
    \centering
    \includegraphics[width=\linewidth]{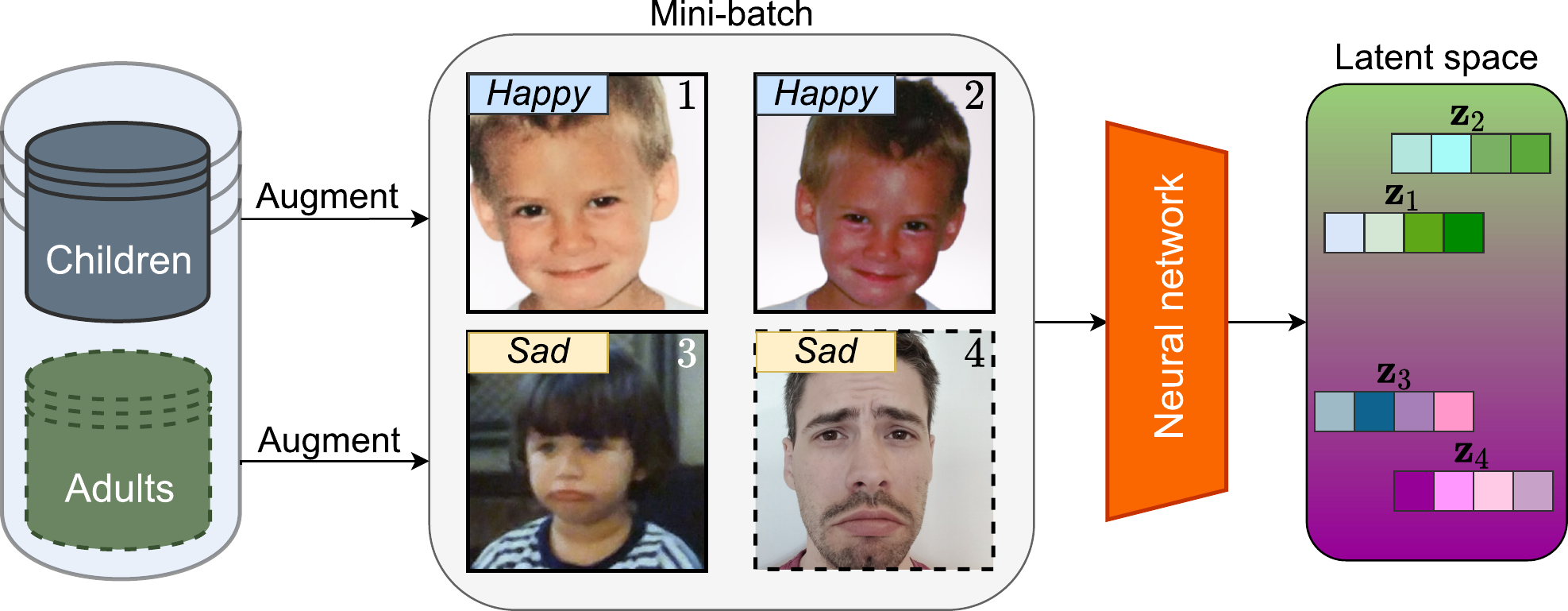}
    \caption{
    Contrastive learning methods train a network to encode data points (faces of children) into a latent representation by which similar images result in spatially-close encodings.
    Normally, this is done using augmentations for images from of a single data set (e.g., children data set, images~1 and 2), with or without label (e.g., as in SupCon and SimCLR, respectively). 
    Here, we explore the effect of introducing data from another distribution (adults data set), by simply adding them to the first data set, or by using them as augmentations for the data from the first data set (e.g, images~3 and 4).
    }
    \label{fig:supcon}
\end{figure}

\Cref{fig:supcon} summarizes, at high level, the contrastive learning methods considered here.
Given a data set (e.g., of children's facial emotions), a neural network is trained with mini-batches that contain augmentations of the original images, to learn to project images that are \emph{similar} into respective latent representations that are \emph{close}.
A label may or may not be present and used by the contrastive learning method.
Different from the traditional use of a single data set (e.g., as would be the case for images 1 and 2 in the mini-batch of \Cref{fig:supcon}), here we juxtapose a second data set (of adults) that shares images that have the same set of (possibly unknown) labels as the first data set (of children), and study how this can be exploited to improve the network's capability to learn a good latent representation.

In the next section, we describe SupCon and SimCLR in more detail.
Next, in~\Cref{sec:approaches} we describe the approaches that we investigate to leverage adult facial expressions.
Details on the adopted data sets and experimental setup are then provided in \Cref{sec:experimental-setup}.

\subsection{Contrastive learning: SimCLR and SupCon}\label{sec:supcon}

We begin by describing the principles behind \cite{chen2020simple}'s SimCLR.
SimCLR does not require labels, and is thus an \emph{unsupervised} or \emph{self-supervised method}.
Let $b$ be the size of a mini-batch. 
SimCLR operates by (i) randomly sampling $b/2$ images from the data set; (ii) generating \emph{two} augmentations for each image; (iii) filling the mini-batch with such augmentations; and (iv) computing the loss:
\begin{equation*}
    \mathcal{L^\text{SimCLR}} = - \sum_{i \in I} \log \frac{\exp \left( \textit{s}(\mathbf{z}_i, \mathbf{z}_{o(i)}) / \tau \right)}{ \sum_{q \in I \setminus \{i\}} \exp\left( \textit{s}(\mathbf{z}_i, \mathbf{z}_q) / \tau \right)  },
\end{equation*}
where $\mathbf{z} \in \mathbb{R}^K$ is a latent representation that is produced by the network, $I$ is the set of indices that identifies the augmentations in the mini-batch, $i$ is the index of an augmentation, $o(i)$ is the \emph{other} augmentation for the same image for which the $i^\textit{th}$ augmentation was made, $\textit{s}$ is a similarity function (here, cosine similarity), and $\tau$ is a hyper-parameter.
Essentially, $\mathcal{L^\text{SimCLR}}$ increases the similarity of latent representations generated for those elements of the mini-batch that are augmentations of the same image.


One can easily see that SimCLR relies on the mini-batch containing mostly (augmentations of) images with different (unknown) labels: if two images are retrieved from the data set that share the same meaning, their respective augmentations will be treated as dissimilar by $\mathcal{L^\text{SimCLR}}$.
Differently from SimCLR, \cite{khosla2020supervised}'s SupCon assumes that labels are available and uses them to tackle this limitation, by:
\begin{equation*}
    \mathcal{L^\text{SupCon}} = \sum_{i \in I} \frac{-1}{|P(i)|}\sum_{p \in P(i)} \log \frac{\exp \left( \textit{s} (\mathbf{z}_i, \mathbf{z}_p) / \tau \right)}{ \sum_{q \in I \setminus \{i\}} \exp\left(\textit{s}(\mathbf{z}_i, \mathbf{z}_q) / \tau \right)  },
\end{equation*}
where $P(i)$ is the set of indices different from $i$ that share the same label of the $i$-th augmentation.
Note that, in principle, SupCon does not need to rely on augmentations, and could rely entirely on the images that are present in the data set.
However,~\cite{khosla2020supervised} propose to use the same procedure that is used for SimCLR, i.e., to craft two augmentations of the same image to populate a mini-batch.

\subsection{Considered approaches}\label{sec:approaches}

We intend to adopt two data sets, namely a \emph{data set A} for which the latent representations are ultimately needed (here, for facial expressions of children); and a \emph{data set B} that acts as a supplement of information (here, facial expressions of adults).
We consider two approaches to use these data sets:
\begin{enumerate}
    \item Simply extend data set A with images from data set B, de facto adopting the union of the two as a new data set.
    Each mini-batch will then contain augmentations originating from both data sets.
    We sample images from the two data sets with equal proportion.
    \item \emph{Explicitly} use augmentations of data points from data set B as if they were augmentations of points of data set A.
    Specifically, to populate a mini-batch, firstly we sample $b/2$ images from data set A; next, for each image from data set A, we insert in the mini-batch \emph{one} augmentation of that image, and \emph{one} augmentation of a random image from data set B, that shares the same label.
\end{enumerate}

\begin{figure}
    \centering
    \includegraphics[width=\linewidth]{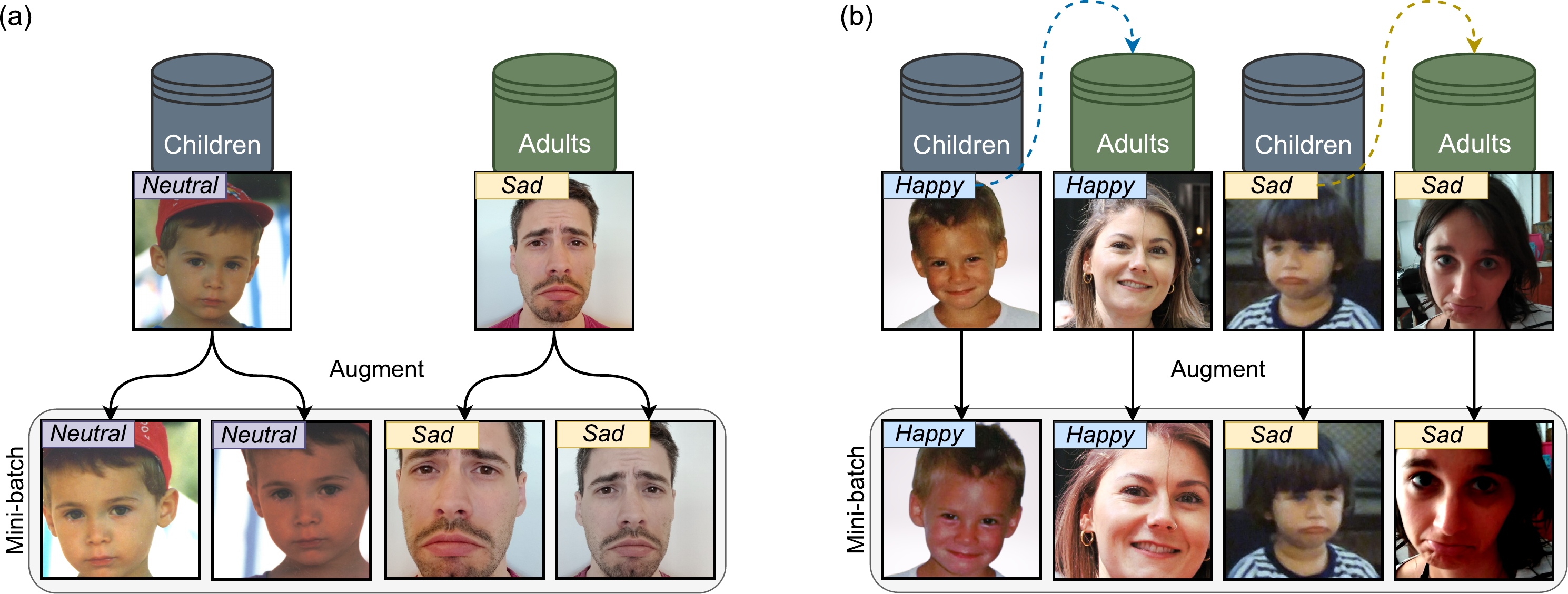}
    \caption{Illustration of the two approaches we consider for populating a mini-batch using two data sets. 
    In the first approach (a), two augmentations are produced for each image taken from a data set. 
    In the second approach (b), the label of the image taken from the first data set (of children) is used to fetch an image from the second data set (of adults).
    An augmentation is produced for each of the two images, and the augmentation of the image from the second data set (of adults) is explicitly used as another augmentation of the image from the first data set (of children).
    }
    \label{fig:supcon-approaches}
\end{figure}

The two approaches are depicted in \Cref{fig:supcon-approaches}.
On the one hand, the two approaches may seem similar because the amount of data points that are taken from data sets A and B are the same in expectation (fifty-fifty in both cases).
On the other hand, the first approach results in mini-batches containing two augmentations of the same image, while the second does not. 
Consequently, one can expect that the amount of information within a mini-batch (informally, how much different the images will be on average) will be larger with the second approach than with the first.
In other words, let $\mathbf{x}^A_i$ be the $i^\textit{th}$ image from data set A, $\mathbf{x}^B_i$ be the $i^\textit{th}$ image from data set B, and $a(\mathbf{x}, \xi)$ be the augmentation produced for image $\mathbf{x}$ using the collection of random variables $\xi$, which represents what random transformations are applied.
Then, for a meaningful similarity function $\sigma$ between images (e.g., one could attempt to measure the proximity of locality sensitive hashing systems such as Apple's \emph{NeuralHash}~\cite{csam2021detection}), it is reasonable to expect that:
\begin{equation*}
    \mathbb{E}_{\xi,i,j} [\sigma(a(\mathbf{x}^A_i, \xi), a(\mathbf{x}^A_j, \xi))] > \mathbb{E}_{\xi,i,j} [\sigma(a(\mathbf{x}^A_i, \xi), a(\mathbf{x}^B_j, \xi))].
\end{equation*}
Since with the second approach mini-batches contain images that are more dissimilar from one another, these mini-batches may carry a richer training signal than those obtained with the first approach.

We remark that the second approach cannot be used with SimCLR, as one needs to know the labels to pick images from data set B for the images of data set A.

\subsection{Experimental setup}\label{sec:experimental-setup}
We set up our experiments according to the settings used by~\cite{khosla2020supervised}, with adaptations for the task and data at hand.
We begin by describing the data sets used and their pre-processing.
Next, we provide details on the possible augmentations, followed by information on the network, the training process, and the validation steps.

\subsubsection{Data sets}
We consider two data sets: the Child Affective Facial Expression (CAFE) data set~\cite{lobue2015child}, and the Facial Expression Recognition challenge (FER) data set~\cite{goodfellow2013challenges}.
CAFE contains $\num{1192}$ pictures of $2$ to $8$ years old-children, in color and portrait format, for the following emotions: anger, disgust, fear, happiness, neutral, sadness, surprise.
FER contains $\num{28000}$ images for the same set of emotions, in gray scale and squared format. 
FER is mostly populated by images of adults, however some pictures of children are also present. 
By visual estimation of a random sample of $1050$ images ($150$ per class), we estimate that approximately $15\%$ of the images in FER represent children up to ten years old (some of whom are younger than the children represented in CAFE, i.e., $< 2$ y.o.).

We pre-process CAFE to have a same picture format of FER, i.e., the top and the bottom of the images is appropriately cut (based on visual inspection) to obtain a $1:1$ width-height ratio, resolution is set to $48$ pixels per dimension, and images are converted to gray scale. 

Throughout the experiments, CAFE is used to train, validate, and test the considered approaches. 
We assess how the results change based on how much data are available for training, by using training splits of $50\%$, $70\%$, and $90\%$; the validation and test sets are created by partitioning the remaining data points in equal amounts.
We use balanced training sets, assigning the number of data points for training by multiplying the training split percentage with the number of data points available in the minority class.
We repeat each experiment ten times due to the stochastic nature of data splitting and neural training.
For FER no splits are used, as it is only used to supply random images of adults during training.

\subsubsection{Neural architecture, training process, and augmentations}

\paragraph{Architecture}
The neural architecture used for contrastive learning (SupCon and SimCLR) is composed of two parts: an \emph{encoder network} and a \emph{projection network}.
The encoder network takes an image $\mathbf{x}$ and produces a latent representation $\mathbf{r} \in \mathbb{R}^{K^\prime}$. 
Like~\cite{khosla2020supervised}, we use a ResNet~\cite{he2016deep} and in particular a gray-scale version of ResNet-18, with $K'=512$.
The projection network is a multi-layer perceptron stacked on top of the encoder network, to map $\mathbf{r}$ to $\mathbf{z} \in \mathbb{R}^{K}$, which is a smaller projection (i.e., $K < K^\prime$).
The contrastive loss is computed upon the representations $\mathbf{z}_1$, \dots, $\mathbf{z}_b$ obtained by iteratively processing (by the encoder network and the projection network) the respective $\mathbf{x}_1, \dots, \mathbf{x}_b$, where $b$ is the size of the mini-batches.
We use $K=128$ and $b=32$.
The projection network is discarded after constrastive training.
To perform classification, a linear classifier is stacked on top of the encoder and trained with a cross-entropy loss.
We refer to the work by~\cite{khosla2020supervised} for more details.

\paragraph{Training and validation settings}
Our experimental settings for training are essentially those used by~\cite{khosla2020supervised}, adapted to the smaller size of the data sets at play here.
In particular, we train for $n_\text{epochs}=250$, as this is abundantly sufficient to make the contrastive losses of SupCon and SimCLR converge in all our experiments\footnote{Training to convergence is often done in contrastive learning because one lacks a classifier (or, for SimCLR, even labels) to perform early stopping.}.
For stochastic gradient descent, we set the initial learning rate to $0.05$, momentum to $0.9$, and learning rate decay to $0.1$ at epochs $n_\text{epochs} - 0.1 \times n_\text{epochs} \times \{3, 2, 1\}$.
The linear layer is then trained for $50$ epochs with similar settings (except for using an initial learning rate of $10^{-4}$), which we found to be sufficient to for the loss to converge.
During these epochs, we record the parameters of the network that lead to the best top-1 validation accuracy, and use those parameters for testing.

We also include a comparison with classic supervised learning training, where a network composed of the encoder and the linear classifier is trained directly with the cross-entropy loss for $n_\text{epochs}=250$.
Again, the best-found validation accuracy is tracked to determine which parameters to use for testing.

\paragraph{Augmentations}\label{sec:augmentations}
We follow~\cite{khosla2020supervised} and adopt cropping of up to $20\%$ of the image; horizontal flipping with $0.5$ probability; and color jittering of $\pm 0.4$ for brightness, contrast, saturation, and $\pm 0.1$ for hue, with $0.8$ probability.

We formulate the use of augmentations in a different way than how typically done in literature.
Normally, one or two augmentations are generated per image on the fly, to enter a mini-batch~\cite{khosla2020supervised,fort2021drawing}.
As a consequence, the longer the training process takes, the more information is fed to the network.
Here, we re-frame this phenomenon in an equivalent setting whereby training has a fixed duration (of $n_\text{epochs}=250$), but the size of the training set varies based on how many augmentations calls are made.
This allows us to directly think in terms of how much data is seen by the network with respect to the size of the training set.
In the results below, we will use the term \emph{augmentation ratio} to refer to the multiplicity by which the original training set is used to craft augmentations.

\section{Results}
We present two types of results:  accuracy (on validation and test sets), and a qualitative comparison using class activation maps.
From now onwards, we use the following acronyms:
SC for \emph{supervised contrastive learning}, i.e., SupCon; UC for \emph{unsupervised contrastive learning}, i.e., SimCLR; and SL for traditional \emph{supervised learning}, i.e., the direct training of the encoder and linear classifier, without the intermediate contrastive training.
Furthermore, we indicate whether the approach is trained on CAFE alone with the notation (c) for ``children''; on CAFE extended with an equal number of random images from FER with (c+a) for ``children plus adults''; and on CAFE using images from FER as augmentations for the images in CAFE with (c$\leftarrow$a) for ``children augmented by adults''.
The latter approach is only possible when using SC, as explained before in \Cref{sec:supcon}.

\subsection{Child emotion recognition accuracy}
\begin{figure*}
    \centering
    \begin{tabular}{cc}
        \includegraphics[width=0.47\linewidth]{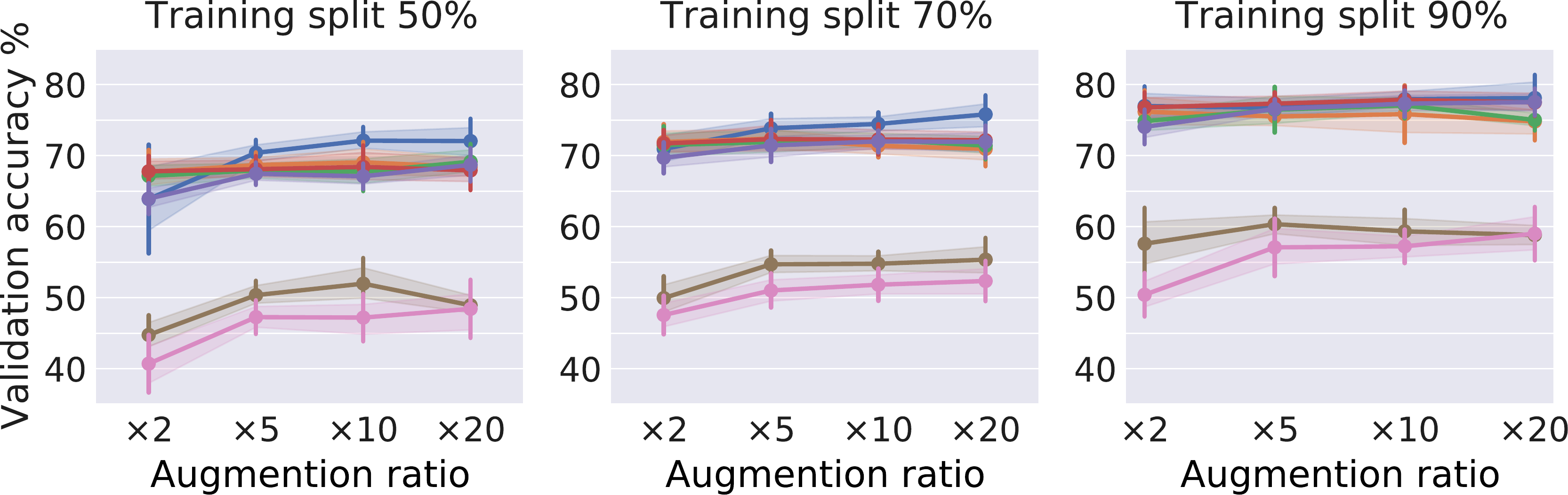} &
        \includegraphics[width=0.47\linewidth]{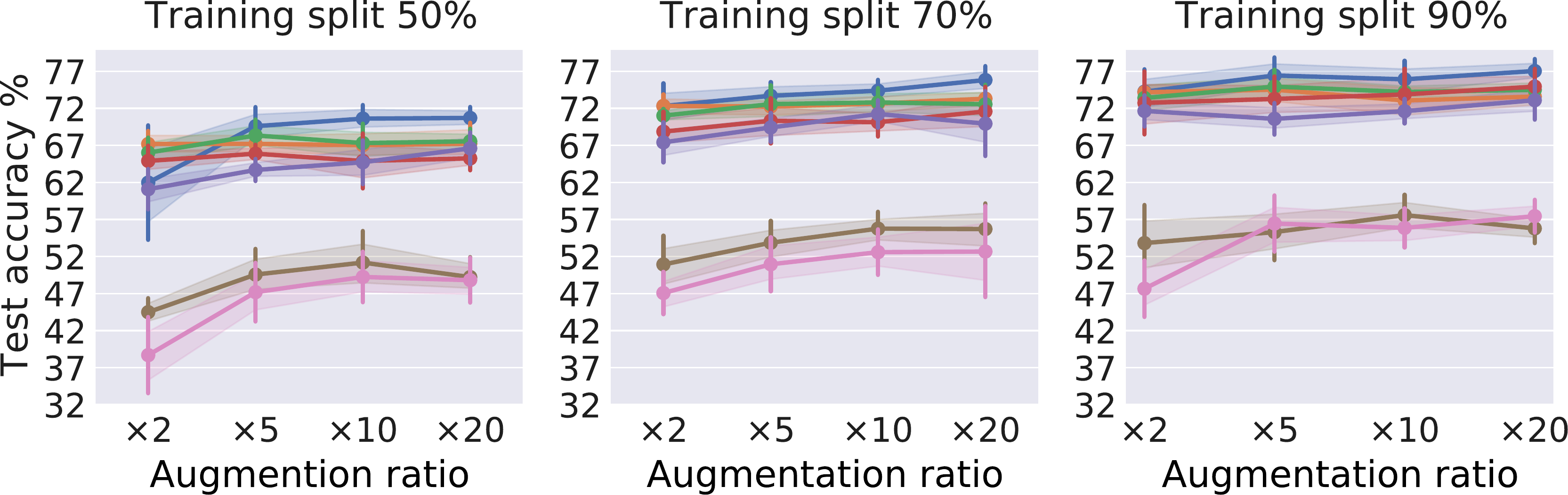} \\
        \includegraphics[width=0.47\linewidth]{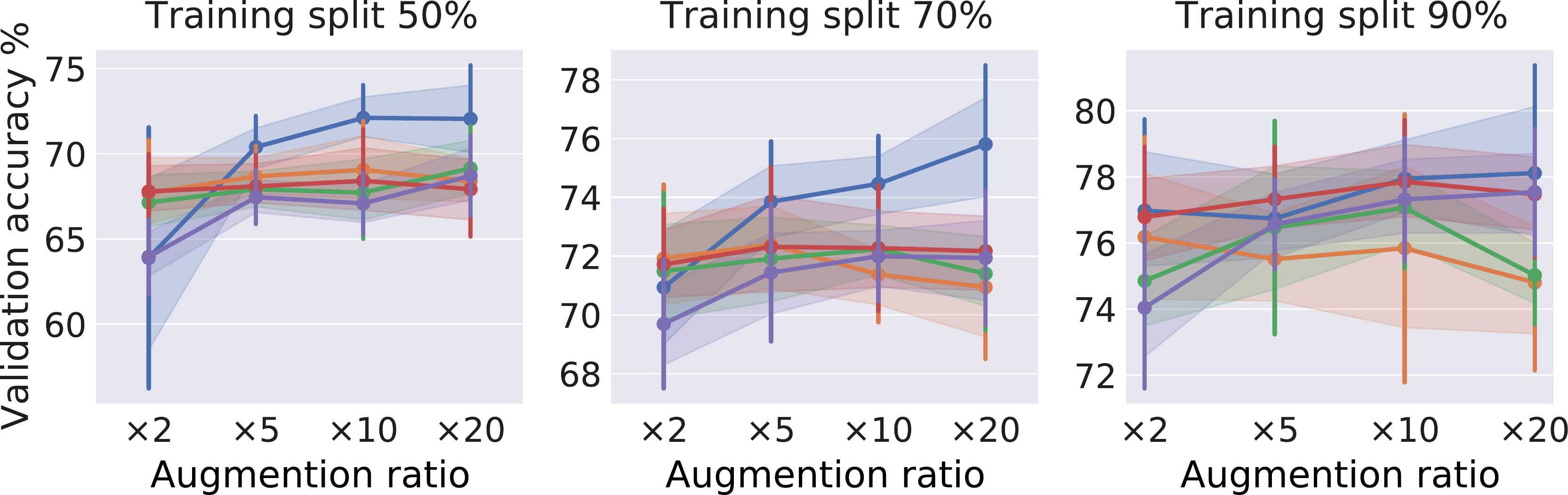} & 
        \includegraphics[width=0.47\linewidth]{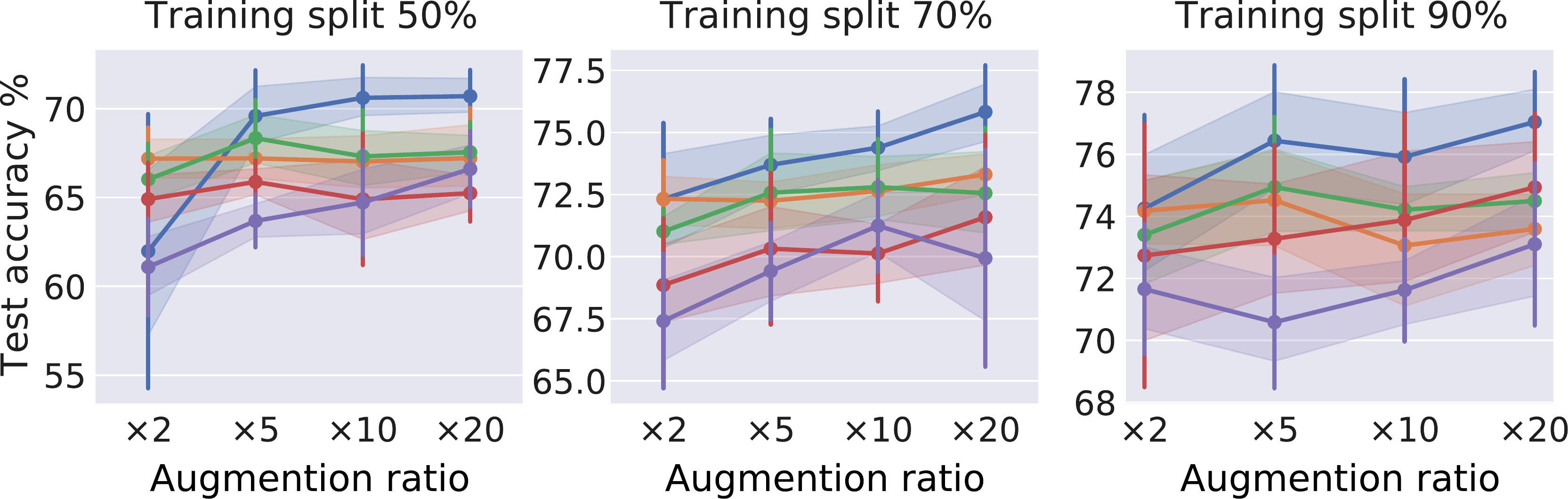}
        \\ 
        \multicolumn{2}{c}{\includegraphics[width=0.67\linewidth]{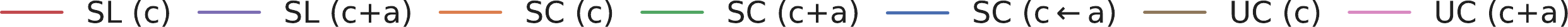}}
    \end{tabular}
    \caption{Validation (left columns) and test (right columns) accuracy for an increasingly larger training set (training split $X\%$) and multiplicities by which augmentations are sampled from the training set (augmentation ratio).
    The panels in the top (bottom) row include (exclude) UC and share (do not share) the same vertical axis.
    Dots are means, shaded areas are $95$\% confidence intervals, and bars are standard deviations, computed for ten repetitions.
    }
    \label{fig:supcon-results-all}
\end{figure*}

Figure~\ref{fig:supcon-results-all} shows the validation and test accuracy for the considered learning approaches, for different sizes of the original training set, and augmentation ratios.

The general trend suggests that using a larger augmentation ratio leads to equal or better validation and test accuracy.
This trend is more pronounced when the training split is small, and less when it is large.
A first evident finding is that the UC approaches perform substantially worse than the others, meaning that the latent representations they learn are poor.
This can be expected because the training data sets are relatively small (UC approaches are normally used on very large, unlabelled data sets) and the number of classes for the problem at hand is also small, hence $\mathcal{L^\text{SimCLR}}$ often puts in contrast augmentations of different images of the same class.

The other approaches, i.e., the variants of SL and SC, perform similarly.
In particular, no evident differences can be observed for SC when CAFE is used alone, i.e., SC (c), or in conjunction with FER, i.e., SC (c+a).
For SL, (c+a) often leads to worse results than (c).
Conversely, the performance obtained by SC (c$\leftarrow$a) is notable, as this approach obtains the best distributions of validation and test accuracies in most cases, except when the training split is large ($90\%$).
There are two (non-mutually exclusive) explanations for this.
First, with a training split of $90\%$, the training data may already be sufficient to learn a latent representation that is (almost) optimal. 
Second, as the validation and test sets are each composed of half the data left over from the training split, the recordings of validation and test accuracy may be noisy.
Lastly, for SC (c) and SC (c+a) with a training split of $90\%$ and the larger augmentation ratios, the best-found validation accuracy drops, which indicates over-fitting of the contrastive training process.

\Cref{tab:summary-test-acc} shows results for the best and second-best approaches in terms of test accuracy, for a $\times 20$ augmentation ratio and across the different training splits. SC (c$\leftarrow$a) is always the best method, whereas the second-best varies. The Mann-Whitney-U test is used to report the $p$-values of the null hypothesis that the best approach is not significantly better than the second-best approach~\cite{mann1947test}.
The improvements in test accuracy vary between $2$ to $3\%$, and are significant with $p\text{-value} < 0.1$ for all cases, and with $p\text{-value} < 0.05$ when the training split is $50\%$ and $70\%$.

\begin{table}[]
    \caption{Mean test results for $\times 20$ augmentation ratios of the best and runner up (R.U.) approaches across training split sizes (Tr.S.).}
    \label{tab:summary-test-acc}
    \centering
    \small
    \begin{tabular}{cl S[table-format=2.1] l S[table-format=2.1] c}
    \toprule
        Tr.S. & Best & {B.~Acc.} & R.U. & {R.U.~acc.} & $p$-val. \\
    \midrule
        50\% & SC (c$\leftarrow$a) & 70.7 & SC (c+a) & 67.6 & $.001$\\
        70\% & SC (c$\leftarrow$a) & 75.8 & SC (c) & 73.3 & $.007$  \\
        90\% & SC (c$\leftarrow$a) & 77.0 & SL (c) & 74.9 & $.061$  \\
    \bottomrule
    \end{tabular}
\end{table}

\begin{figure}[h]
    \centering
    \includegraphics[width=\linewidth]{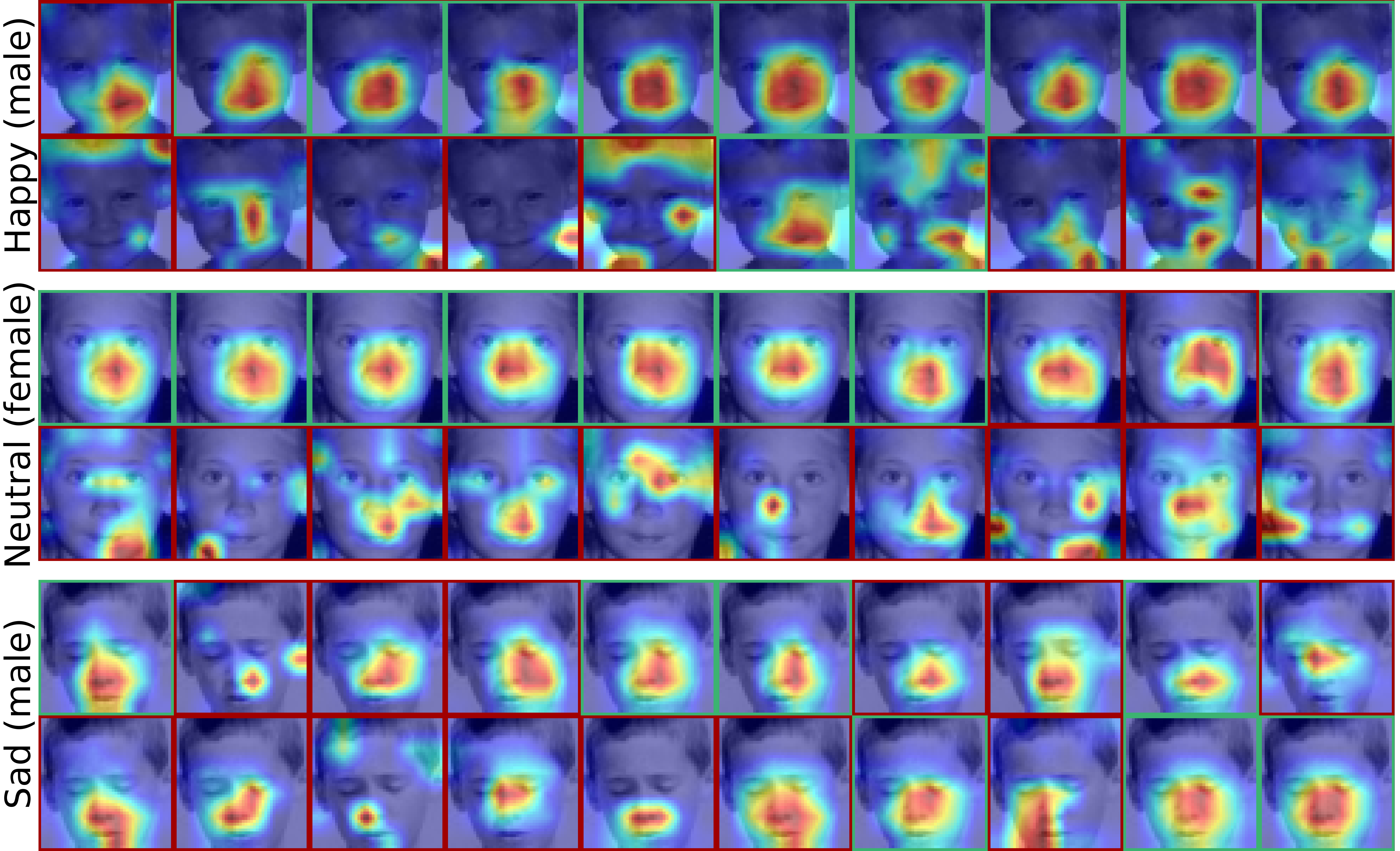}
    \caption{Class activation maps for three children's facial expression images. 
    The more the color of a pixel is red (as opposed to blue), the more the pixel is deemed to be responsible for the activation of the true class (reported above each pair of rows).
    For a pair of rows, the top row shows the class activation maps obtained by the ten networks trained with SC (c$\leftarrow$a), while the bottom row shows the ten networks trained with SC (c).
    When the prediction of a network is incorrect, the border of the corresponding image is red (instead of green), and the class activation map often appears to be unreasonable.
    }
    \label{fig:saliency-maps}
\end{figure}

\subsection{Qualitative analysis with class activation maps}
We provide additional insights by comparing the class activation maps that are obtained by requiring the networks to identify the areas of the image that prompt its prediction to provide the \emph{correct} label.
We consider SC (c$\leftarrow$a) (the best performing approach that makes the best use of adult faces) and SC (c) (the respective approach that uses only children's faces) for a training split of $70\%$ and augmentation ratio of $5$ (an intermediate training configuration).
Because the images of CAFE are protected by copyright, we use images of the authors and their family (similarly to previous figures). 
Note that these images can be considered to be out-of-distribution because they do not belong to CAFE (nor FER, for SC (c$\leftarrow$a)).
Class activation maps are computed with the implementation by~\cite{jacobgilpytorchcam} of Grad-CAM~\cite{selvaraju2017grad}.

\Cref{fig:saliency-maps} shows the class activation maps obtained for the true class with the networks trained by ten repetitions of the training process (columns). 
Images for which the network predicts the correct (respectively, incorrect) class have a green (red) border.
The ten networks trained with SC (c$\leftarrow$a) are, on average, more accurate than those trained with SC (c).
Interestingly, the area highlighted for correct predictions (green border images) corresponding to different networks from the same approach (a same row) can differ, due to the randomness of the data splitting and training process. 
For example, sometimes the eyes are deemed to be important, and sometimes they are not.
When the prediction is incorrect (red border images), the pixels that are deemed to be responsible by the class activation map tend to be scattered around the image, at times on parts that can be considered not helpful to predict the emotion (e.g., part of the hair or the chin). 
For SC (c), one of such ``unreasonable'' maps is also produced for a correct classification of the happy boy, on the $7^\textit{th}$ column.

\section{Discussion}
In the introduction of this work, we made two hypotheses.
First, that training a neural network with more and more augmentations of the same data has diminishing returns.
Our results confirm this hypothesis. 
In fact, this finding can also be observed in the recent work by \cite{fort2021drawing} for much bigger data sets than the one used here (CAFE), namely CIFAR-100 and ImageNet~\cite{krizhevsky2009learning,deng2009imagenet}.
However, our results also show that diminishing returns are less pronounced when the training data set, i.e., images of children, is supplemented with extra data, i.e., images of adults, provided that the training set is relatively small and an appropriate learning approach is used.
Thus, in settings where the data for the problem at hand is scarce (e.g., as often the case in pediatric medicine), augmentations can only help up to some point. 
However, integrating a different but related data source can bring additional improvements.

Our second hypothesis was that using adult data can be beneficial for recognizing the emotions of children.
While the best approach, SC~(c$\leftarrow$a), does in fact make use of adult data, that was not always the case for the second-best approach, indicating that \emph{how} the adult data are used \emph{matters}.
Thus, our results support this hypothesis only in part.

Related to this work, \cite{fort2021drawing} proposed to use multiple augmentations of a few images when populating a mini-batch, to diminish the variance of the training process.
They found that using larger augmentation multiplicities can improve both training speed and generalization performance, when dealing with relatively large data sets (compared to CAFE), namely CIFAR-100 and ImageNet.
Interestingly, such an approach is more similar to our SC~(c) and SC~(c+a), which use two augmentations of the same image per mini-batch, than to SC~(c$\leftarrow$a), which uses a single augmentation of an image per mini-batch (see \Cref{fig:supcon-approaches}).
Yet, SC~(c$\leftarrow$a) was the best-performing approach in our experiments: this might seem counter-intuitive.
Actually, the results by \cite{fort2021drawing} in regimes where over-fitting is likely (i.e., when the number of training epochs is relatively large with respect to the size of the training set) show that increasing the multiplicity of augmentations of the same image increases the speed by which generalization performance drops because of over-fitting.
As CAFE is approximately 50 and \num{1000} times smaller than CIFAR-100 and ImageNet, respectively, CAFE is easier to over-fit.
In this light, our results suggest that for small data sets such as those concerning children, using multiple augmentations of a few images may not be the best approach. 
Rather, one may wish to enrich the informational content of mini-batches (and thus variance in the training process).


\section{Conclusion}
We compared contrastive and supervised learning approaches for training deep neural networks for the problem of facial emotion recognition in children, when the available data are scarce.
We proposed an approach that uses supervised contrastive learning to use abundantly-available data of adults as augmentations for data of children, and found this approach to perform better than baseline approaches, especially when the training set is particularly small. 
Our approach obtained $+2$ to $3\%$ test accuracy compared to the second-best method.
We further showed examples of class activation maps, which are useful tools to explain the behavior of black-box models like deep neural networks. 
This provided additional indication that appropriately incorporating adult data can lead to learning better representations than solely relying on data of children.
This work will hopefully serve as an inspiration for other applications in pediatric care where data is scarce but other sources may be exploited to learn good representations.

\section*{Acknowledgments}
The authors thank Jamal Toutouh for insightful discussions. 
This work is part of the project \emph{Improving Childhood Cancer Care when Parents Cannot be There --
Reducing Medical Traumatic Stress in Childhood Cancer Patients by Bonding with a Robot Companion} (with project number \#15198) of the research program \emph{Technology for Oncology}, which is financed by the Dutch Research Council (NWO), the Dutch Cancer Society (KWF), TKI Life Sciences \& Health, Asolutions, Brocacef, Cancer Health Coach, and Focal Meditech. 
The research consortium consists of Centrum Wiskunde \& Informatica, Delft University of Technology, the Amsterdam University Medical Centers location AMC, and the Princess M\'axima Center for Pediatric Oncology.

\bibliographystyle{named}
\bibliography{ijcai22}

\end{document}